\documentclass[letterpaper, 10 pt, conference]{ieeeconf} 

\IEEEoverridecommandlockouts                              
\usepackage{graphics} 
\usepackage{epsfig} 
\usepackage{mathptmx} 
\usepackage{times} 
\usepackage{amsmath} 

\usepackage{amssymb}  
\usepackage{siunitx}
\usepackage[euler]{textgreek}
\usepackage{float}

\begin{document}
    \title{\Large \bf
    A light- and heat-seeking vine-inspired robot with material-level responsiveness
    }
  \author{
      Shivani Deglurkar$^{1*}$, Charles Xiao$^{1*}$, Luke Gockowski$^{1,2}$, Megan T. Valentine$^{1}$, and Elliot W. Hawkes$^{1**}$
      \thanks{This work was supported in part by NSF Grants CMMI-1944816 and EFMA-1935327. The work of C. Xiao was supported by the NSF Graduate Research Fellowship Program (2139319).}
  \thanks{$^1$Department of Mechanical Engineering, University of California, Santa Barbara, CA 93106.}
    \thanks{$^2$Accenture Labs, San Francisco, CA 94105}
   \thanks{$^*$ Equal contribution}
   \thanks{$^{**}$ Email: ewhawkes@ucsb.edu}
   \thanks{This work has been submitted to the IEEE for possible publication. Copyright may be transferred without notice, after which this version may no longer be accessible.}
}


\maketitle

\begin{abstract}
The fields of soft and bio-inspired robotics promise to imbue synthetic systems with capabilities found in the natural world. However, many of these biological capabilities are yet to be realized. For example, vines in nature direct growth via \textcolor{black}{localized responses embedded in the cells of vine body, allowing an organism without a central brain to successfully} search for resources (e.g., light). Yet to date, vine-inspired robots have yet to show such \textcolor{black}{localized embedded responsiveness}. Here we present a vine-inspired robotic device with material-level \textcolor{black}{responses} embedded in its skin and capable of “growing” and steering toward \textcolor{black}{either a} light \textcolor{black}{or heat} stimulus. We present basic modeling of the concept, design details, and experimental results showing its behavior in response to infrared (IR) \textcolor{black}{and visible} light. Our simple design concept advances the capabilities of bio-inspired robots and lays the foundation for future “growing” robots that \textcolor{black}{are capable of seeking light or heat, yet are extremely simple and low-cost}. 
\textcolor{black}{Potential applications include solar tracking, and in the future, firefighting smoldering fires. We envision using similar robots to find hot spots in hard-to-access environments, allowing us to put out potentially long-burning fires faster.}

\end{abstract}

\section{Introduction}
Over the past decades, bio-inspired robotics and then soft robotics have gained interest, owing partly to their ability to easily adapt to changing environments without complex mechanisms \textcolor{black}{\cite{laschi2016soft}.
\textcolor{black}{Indeed, soft robotics has shown an exponential rate of growth of publications over the last two decades \cite{hawkes2021hard}.}}
However, many intriguing behaviors from the natural world are \textcolor{black}{not} yet realized in these compliant robots. 

One particularly compelling example is from the plant kingdom: tropisms, or directed motility along a gradient, \textcolor{black}{such as sunlight or soil moisture \cite{mazzolai2018growth, forde2001nutritional}. 
This modality is compelling for multiple reasons. First, it requires no complex, centralized controller; instead the response is localized in each growing body. Second, it is scalable;} hundreds or even thousands of individual roots can simultaneously steer. \textcolor{black}{Third}, \textcolor{black}{it is robust;} if some of the vines or roots are damaged or removed, the others remain fully functional. 

 \begin{figure}[h!]
  \begin{center}
  \includegraphics[width=0.9\columnwidth]{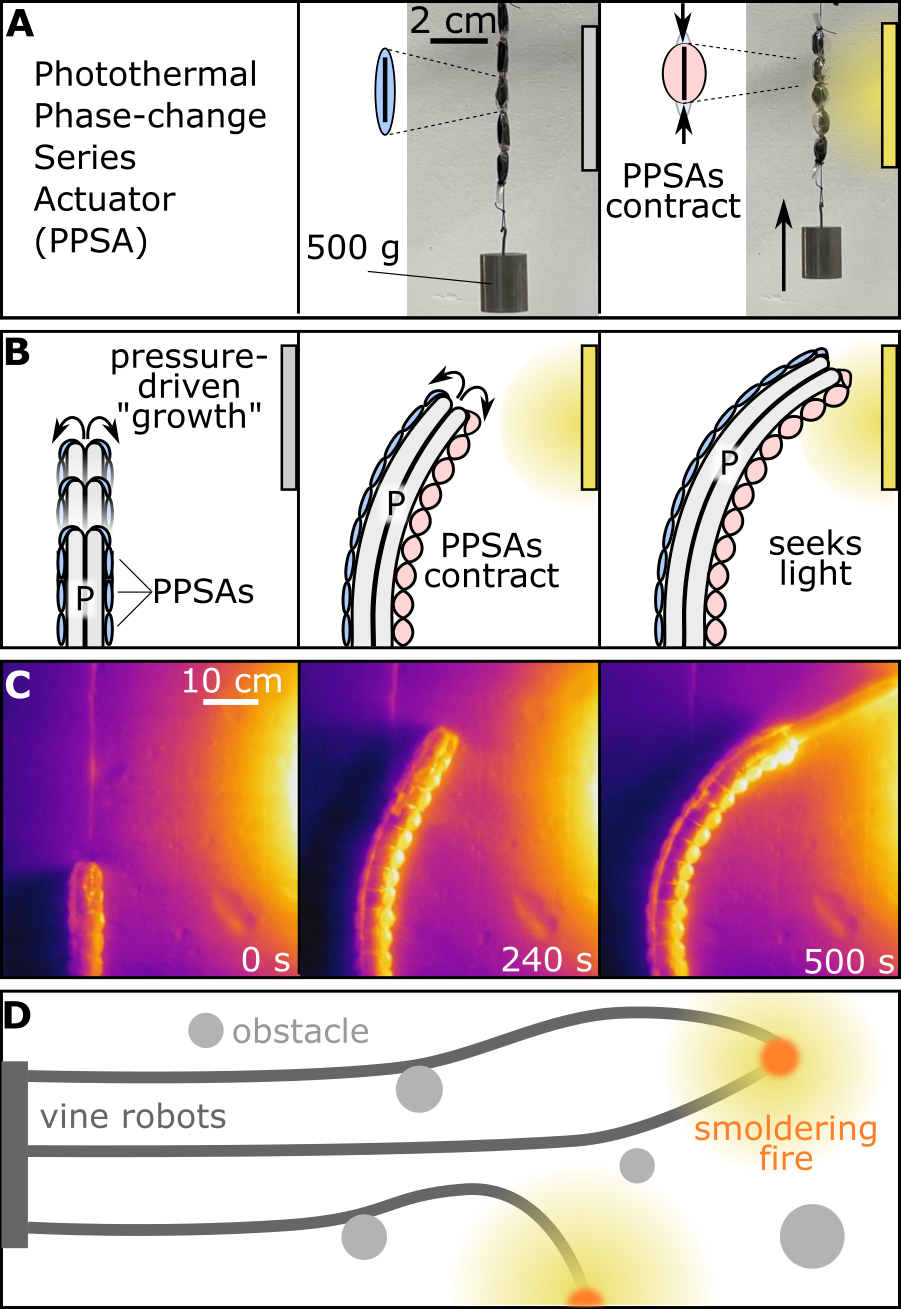}
  \caption{\textbf{A:} We introduce Photothermal Phase-change Series Actuators (PPSAs), which \textcolor{black}{comprise a photoabsorber (black), which converts light to heat, and} low-boiling point Novec 7000. Under light irradiation, the PPSAs absorb light, heat up, inflate with vapor, and contract to lift a mass. \textbf{B:} The PPSA can be incorporated onto a vine-inspired robot body. When the robot is inflated to pressure, $P$, it ``grows'' from the tip by unfurling material from inside. At the same time, the PPSAs on the side exposed to light contract, \textcolor{black}{steering} to the stimulus. \textbf{C:} 
  IR camera image, showing PPSAs heating on the side closest to the light. Temperatures range from approximately 20 (yellow) to \SI{40}{\celsius} (dark purple). 
  \textcolor{black}{\textbf{D:} These robots could eventually be used to seek smoldering fires and deliver fire suppressant.} 
  }
  \label{fig:first_fig}
  \end{center}
  \vspace{-7mm} 
\end{figure}

Various vine- and root-inspired robots have been proposed that extend from their tips to “grow,” 
\textcolor{black}{including} “vine robots,” a class of soft robot made of an inverted, flexible, thin-walled pneumatic tube that everts when pressurized, lengthening from its tip \cite{mishima2003development,invendo2006everting,tsukagoshi2011smooth,tsukagoshi2011tip,sadeghi2013robotic,hawkes2017soft}. 
Because this lengthening involves no relative movement of the body with respect to the environment, the robot can reliably extend through constrained environments, even when the properties of the path and obstacles are unknown, \textcolor{black}{e.g., \cite{hawkes2017soft,greer2018obstacle}. 
However, none of these has demonstrated localized embedded responses to control growth direction.}
\textcolor{black}{Instead, many vine robots utilize} human-controlled teleoperation (e.g., \cite{CoadRAM2019}), \textcolor{black}{or incorporate} autonomous control, 
(e.g., \cite{hawkes2017soft}). Another method \textcolor{black}{incorporates} tropisms,
sensing stimuli such as gravity and moisture, yet rely on a centralized framework for computing the response to sensors and sending signals to actuators \cite{sadeghi2016plant}.

\textcolor{black}{Although all of these methods of controlling the steering direction of growing robots work, they require electronic sensors attached to the robot's body and either a human operator or controller to implement steering. Further, each degree of freedom requires an actuator and a connection to pass a signal.}
\textcolor{black}{Such} steering mechanisms for \textcolor{black}{previous} vine robots include motorized pull tendons (e.g., \cite{wang2020dexterous}), pneumatically-controlled latches (e.g., \cite{hawkes2017soft}), and artificial muscles (e.g., \cite{CoadRAM2019,greer2018soft,selvaggio2020obstacle}).

A promising alternative is a “material-level” scheme that leverages the intrinsic materials properties of the soft robot body. \textcolor{black}{Broadly speaking, this concept has been demonstrated in growing robots that passively deform in response to obstacles to navigate their environment \cite{greer2018obstacle,sadeghi2020passive}}. Outside of growing robots, the capabilities of a number of functional materials to respond to external stimuli (such as light, heat, electromagnetic fields, or chemical conditions) have been demonstrated \cite{shen2020stimuli, zhao2021stimuli}. 
Among these, light and heat are particularly promising due to \textcolor{black}{recent work on} photothermal actuators \cite{photothermalRef1,photothermalRef2, hiraki2020laser}.

In this work, we present a vine-like robot that uses material-level \textcolor{black}{responses} embedded directly in the skin to \textcolor{black}{seek light and heat} (Fig. \ref{fig:first_fig}). These many local \textcolor{black}{embedded responses} are achieved using \textcolor{black}{a photoabsorber} in low-boiling-point fluid that is encased in many small, individual pouches along the sides of an everting soft robot body. We call this new actuator a Photothermal Phase-change Series Actuator (PPSA).
Due to the non-uniformity of light flux around a source, the PPSAs located closest to the source will actuate first, shortening that portion of the robot, and steering it toward the source. 

The primary contribution of this work is advancing the state of the art in vine- and root-inspired “growing” robots by introducing local \textcolor{black}{embedded steering responses}. Secondary contributions include the introduction of PPSAs and a method for integration into “vine” robots, as well as a model of the behavior of the resulting devices. 
\textcolor{black}{We envision using them for solar tracking and in the future, helping put out smoldering fires.}




\section{Device Design}
\label{sec:Design}

\subsection{Robot Design}
We have created a robot that steers toward a light \textcolor{black}{or heat} source as it extends, or ``grows,'' based only on the local \textcolor{black}{responsiveness of the} PPSAs.
The robot has a central spine made of a pneumatically pressurized LDPE tube (\SI{3.2}{\centi\meter} diameter) that provides structure and enables tip-based growth, flanked by two rows of many individually actuated PPSAs (Fig. \ref{fig:first_fig}). 
Left-right steering is achieved by differential inflation of the PPSAs. During growth, the PPSAs  most exposed to the source will most absorb light, increase in temperature, and contract. These PPSAs will be the ones on the side of the robot facing the source and at the position along the robot that is nearest the \textcolor{black}{source}.
The inclusion of an internal photoabsorber (described below) provides a means of “shading” the opposing-side PPSAs, and the pneumatic LDPE spine offers insulation between the two rows of PPSAs--enhancing this differential effect. 
\textcolor{black}{In equilibrium, the robot will bend such that it points at the source, with both sides having equal light exposure.}

\subsection{Photothermal Phase-change Series Actuator (PPSA)}
\label{sec:design}

We introduce the PPSA, a self-contained, untethered, soft photothermal actuator arranged in series. \textcolor{black}{It responds to electromagnetic radiation, or light, because its photoabsorber absorbs photons in a wide range of frequencies (including IR and visible) and produces heat. The heat then boils a phase-change} liquid, increasing pressure inside the actuator, causing radial swelling and length contraction.
\textcolor{black}{Alternatively, the PPSA also responds directly to heat (e.g., through conduction from a hot surface), bypassing the photoabsorber.}
The design builds on previous work using electrically heated \cite{nakahara2017electric} and laser-heated \cite{hiraki2020laser} phase-change liquid in a single flat plastic pouch. \textcolor{black}{However, for vine robots, the actuator must fold into a thin, cylindrical profile to pack into the body of the vine robot before everting from the tip, and produce high strain after everting to curve the body of the vine robot. Accordingly, we designed the PPSA as a pleated, high-strain pouch, inspired by previous pneumatic actuators} \cite{daerden2001concept,daerden1999conception,greer2017series}.
In contrast \textcolor{black}{to these actuators which each need a supply of compressed air}, we isolate the inflation of each PPSA in the series by sealing it individually, thereby enabling localized actuator response. We select Novec 7000 as a working fluid due to its low boiling point (i.e., \SI{34}{\celsius} at atmospheric pressure), non-toxicity, and compatibility with a broad selection of polymer films, as has been done in previous liquid-to-gas \textcolor{black}{actuators. For the skin of the PPSA, we use Mylar because of its low gas permeability and mechanical robustness.} 

 \begin{figure}
  \begin{center}
  \vspace{2mm}
  \includegraphics[width=1\columnwidth]{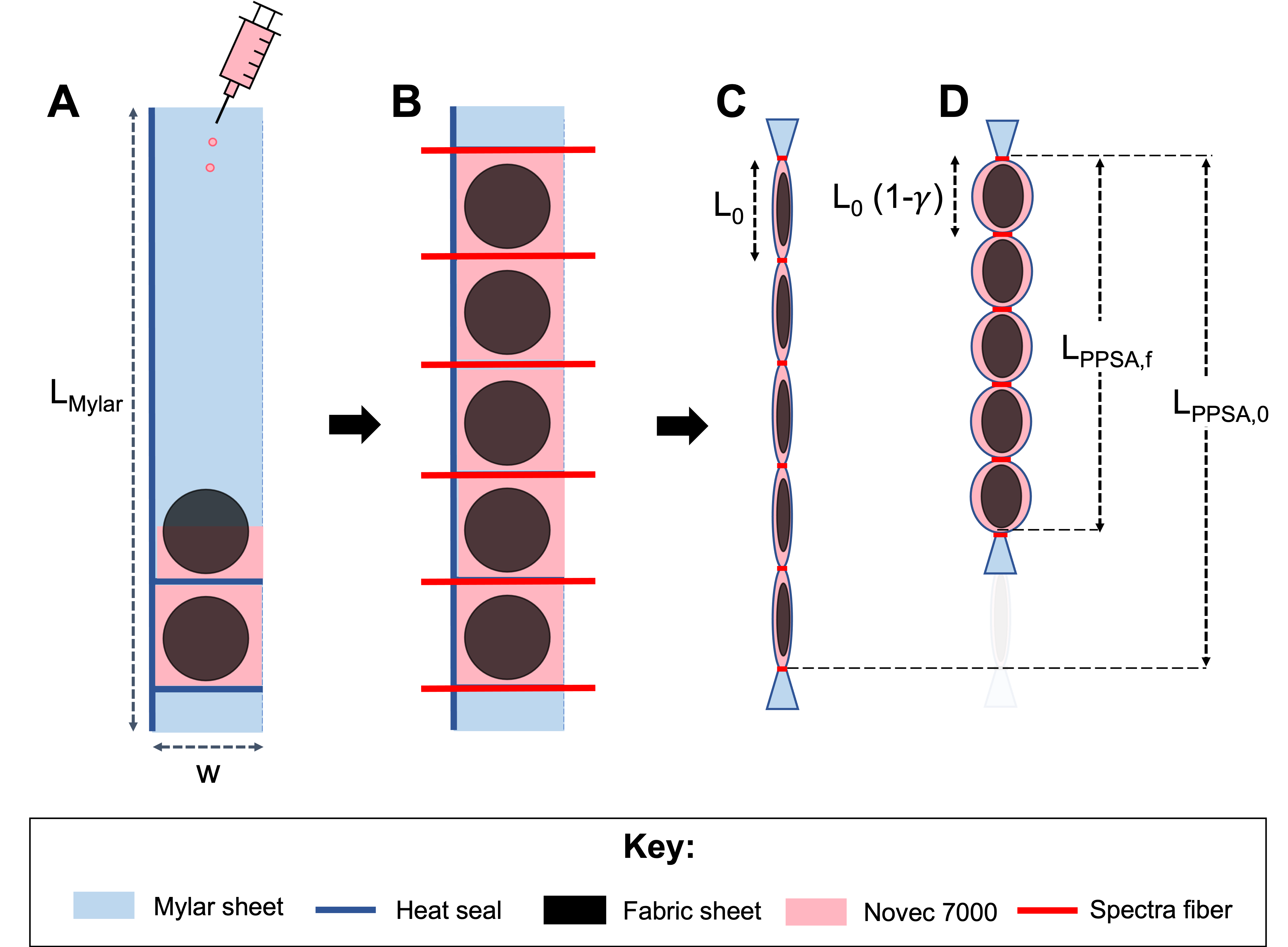}
  \caption{Schematic of fabrication steps for a 5-pouch PPSA. \textbf{A:} Cut Mylar to a \SI{30}{\centi\meter} long ($L_{Mylar}$) by \SI{9} {\centi\meter} wide strip. \textcolor{black}{Fold in half, length-wise, and heat seal} along the length. Create a second heat seal perpendicular to the length and \SI{2.5} {\centi\meter} from the edge to form a \SI{4.5} {\centi\meter} wide ($W$) chamber. To create a pouch, place \SI{3.8} {\centi\meter} diameter circle of absorber into sealed chamber; inject \SI{1.5} {\milli\liter} Novec 7000. Heat seal  further down the length, \SI{4.5} {\centi\meter} from previous pouch. \textbf{B:} Repeat process of creating a pouch for five pouches. \textbf{C:} Tie off pouches using spectra fiber. \textbf{D:} Inflated pouches show contraction.
  }
  \label{fig:fabrication}
  \end{center}
  \vspace{-1mm}
\end{figure}

\textcolor{black}{To choose an absorber configuration,} we measured the time for a single PPSA pouch to transition from fully deflated to fully inflated when fixed to a polystyrene foam block positioned \SI{76}{\centi\meter} away from the \textcolor{black}{IR light}. Each PPSA pouch (\SI{7.0}{\centi\meter} $\times$  \SI{3.7}{\centi\meter}) was heat-sealed with equal volumes of Novec 7000 inside. Four conditions were tested: (A) clear Mylar pouch containing a black nonwoven microfiber fabric sheet (EonTex NW170-PI-20), (B) clear Mylar pouch with the microfiber sheet adhered to the surface using double-sided tape, (C) aluminized Mylar pouch with a microfiber fabric sheet adhered to its surface, (D) aluminized Mylar pouch with the surface coated with matte black spraypaint (Krylon Ultra Flat Camouflage). The response times of conditions A, B, C, and D were 95, 270, 105, and 150 seconds, respectively. The fastest configuration, clear Mylar containing an internal microfiber fabric sheet (condition A), was selected for the design of our PPSA.

\section{Fabrication}

The PPSAs were constructed from a rectangular strip of clear Mylar folded onto itself length-wise, and heat sealed along its length \textcolor{black}{with an impulse heat sealer for 2 seconds (Uline, 450W)}. Then, a perpendicular heat seal created a chamber. Next, a circle of microfiber fabric sheet was slipped in from the unsealed edge, Novec 7000 was injected, and the pouch enclosed with a heat seal (Fig. \ref{fig:fabrication}A). This process was repeated to produce 5 flat pouches in series. Finally, Spectra fiber (\textit{Power Pro} 80 lb test) was tied onto the four seals separating each pouch and at the two ends (Fig. \ref{fig:fabrication}B). \textcolor{black}{The Spectra fibers help maintain the radius of the nodes separating PPSAs.}
This changes the geometry from flat pouches to a pleated, radially symmetric pouch (Fig. \ref{fig:fabrication}C, D).
Constraining the actuator geometry in this way has been demonstrated as a method to achieve higher contraction ratios compared to flat pouches \cite{greer2017series}.  
\textcolor{black}{Multiple 5-pouch} sets of PPSAs were adhered to opposing sides of the LDPE spine using double-sided tape. Lastly, the robot body was inverted and pressurized. \textcolor{black}{When the PPSAs are inside the body, the internal pressure helps maintain them in a tightly folded configuration.}

\section{Semi-empirical Model}
\label{sec:model}
The goal of the model is to predict the pose of the robot for a given light flux at steady state. \textcolor{black}{This can allow a user to predict whether a target light or heat source would be reached by a robot, given a starting position and orientation (see Fig. \ref{fig:HitTest})}. To do this, we use a semi-empirical model, combining an empirical component and an analytical component. 
For the empirical component of the model, we use experimental data to determine the relationship between light flux and contraction ratio of the PPSA.
For the analytical component of the model, we use a static pose model to relate the contraction ratio of the PPSAs to the pose of the robot at equilibrium. 
\textcolor{black}{This model applies to robots with low external loading (e.g., no significant forces from obstacles) and heat transfer conditions similar to  those of experiments reported in Fig. \ref{fig:SimpleEnvironment}}.

\subsection{Empirical Component: Light Flux to Contraction Ratio}

To determine the empirical relationship between flux and contraction ratio, we first measured the flux as a function of distance from the light source (i.e., infrared lamp) and then measured the contraction ratio as a function of distance from the light. By combining the results from these two experiments, we infer the flux-contraction ratio relation.  


\subsubsection{Flux versus Distance from Light}
\begin{figure}[tb]
    \centering
    
    \vspace{2mm}
    \includegraphics[width=1.0\columnwidth]{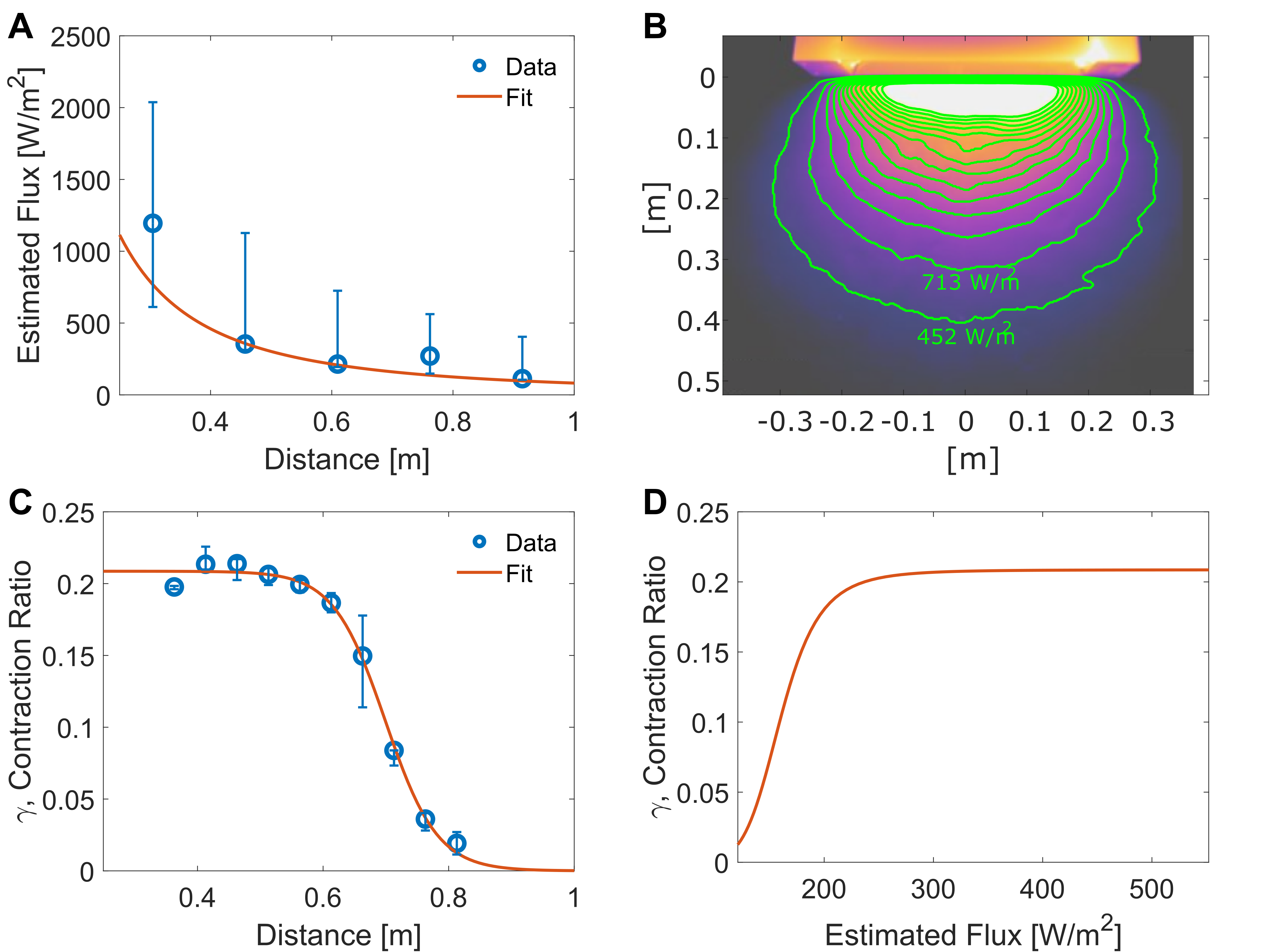}
    \caption{The empirical component of the model is built from flux versus distance data combined with contraction ratio versus distance data. \textbf{A:} First, we measure flux as a function of distance from the light source. The circles are the experimentally measured points and the solid line is an exponential fit to the data. \textbf{B:} IR image of the temperature produced by lamp on a polystyrene foam floor. The contours represent isotherms, which we interpret as isoflux lines. \textbf{C:} Second, we measure contraction ratio as a function of distance from the light source, for a fixed force level of \SI{5}{\newton}. \textbf{D:} Finally, we assemble these two relationships to create a representation of the relationship between contraction ratio and flux.}
    \vspace{-2mm}
    \label{fig:SimpleEnvironment}
\end{figure}

To experimentally determine the relationship between flux and distance from the light source, we  measured the equilibrium temperature of a plate at varying distances.
For the light source, we used an infrared lamp (Infratech W-7512 SS) with an approximate peak emission wavelength of \SI{2.3}{\micro\meter}. 
For the plate, we used a black-painted aluminum plate ($\sim$\SI{1.6}{\centi\meter} $\times$ \SI{1.6}{\centi\meter}). At all separation distances, the center of the plate was placed ~\SI{2}{\centi\meter} above the polystyrene floor, which is at a height that is approximately level with PPSAs on the robot. At each position, we measured and recorded the equilibrium temperature, $T_{plate}$ with a K-type thermocouple (ThermoWorks K-36X) attached to the back side of the plate. For simplicity, we assumed the plate is a blackbody, which is reasonable since the paint has high emissivity. \textcolor{black}{The estimated flux, $Q$, has the form $Q_{light}=\sigma T_{plate}^4-h(T_{plate}-T_{amb})$, where $\sigma$, $h$, and $T_{amb}$ are the Stefan-Boltzman constant, convection coefficient, and ambient temperature, respectively. \textcolor{black}{The fit has the form of $Q=c_1/x^{c_{2}}$, where $Q$ is the estimated flux, $c_1$ is \SI{81.922}{\watt\meter^{-0.117}}, $x$ is distance, and $c_2$ is \SI{1.883}{}}.  As expected, the flux nearly obeys the inverse square law (Fig. \ref{fig:SimpleEnvironment}A). Re-radiation from the ground and the lamp shape largely explains the deviation from the inverse square law.} Using IR imaging (FLIR E60) we determined the spatial distribution of flux around the lamp (Fig. \ref{fig:SimpleEnvironment}B). We assume isotherms are isofluxes.

We attribute the primary uncertainty of Fig. \ref{fig:SimpleEnvironment}A to the uncertainty of the convection model. 
\textcolor{black}{The error bars shows the estimate for flux for different convection coefficients. Free convection coefficients, $h$, typically varies from 2 to \SI{25}{\watt/\meter^2/\celsius} (Table 1.1 of \cite{bergman2011fundamentals}). For our estimates, we use the vertical flat plate convection coefficient model (Eq. 9.27 of \cite{bergman2011fundamentals}), because our boundary conditions are close to those of the model. Eq. 9.27 is an empirical correlation that relates the convection coefficient to temperature and geometric and fluid parameters. Determining the exact coefficients is beyond the scope of this paper.}

  \begin{figure}
  \begin{center}
  \vspace{2mm}\includegraphics[width=1\columnwidth]{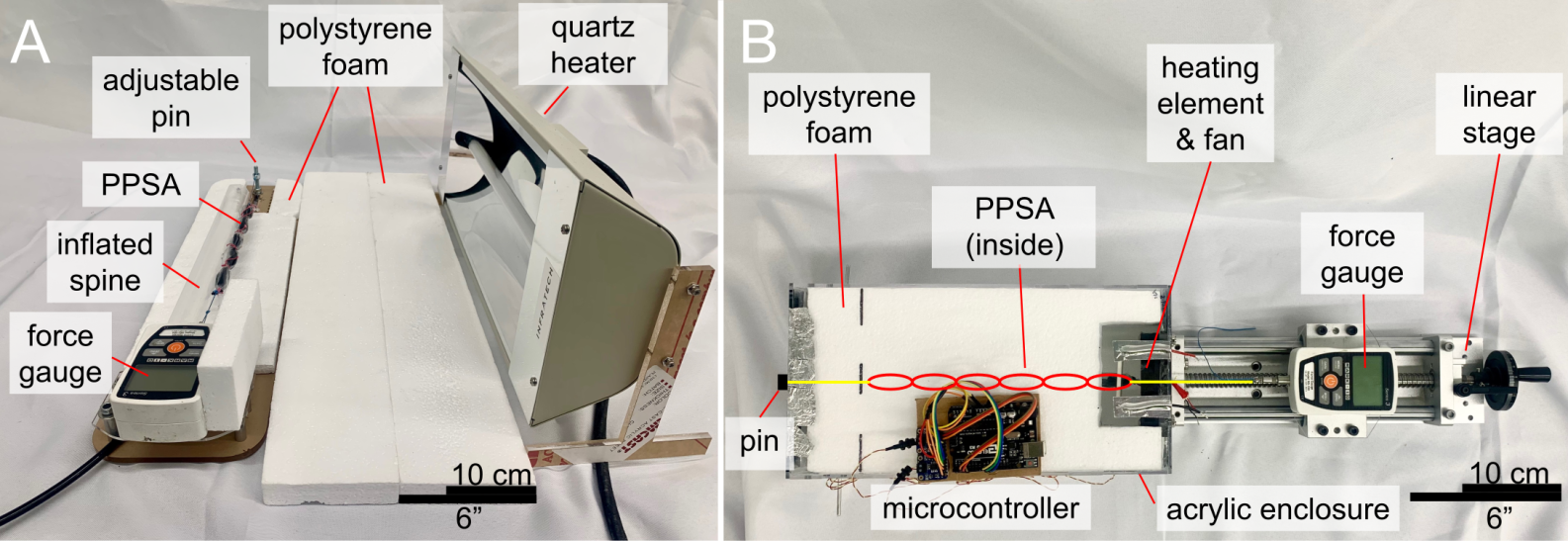}
  \caption{Images of the testing setups. \textbf{A:} Setup for measuring the relationship between flux and the distance from the IR source (Fig. \ref{fig:SimpleEnvironment}). \textbf{B:} Temperature-controlled setup for measuring the force-contraction ratio relationship at different temperatures (Fig. \ref{fig:ForceStrain}).
  }
  \label{fig:outbox_schematic}
  \end{center}
  \vspace{-5mm}
\end{figure}

\subsubsection{Contraction Ratio versus Distance from Light}
\label{sec:Contraction Ratio vs. Flux}
Next, we measured the contraction ratio at different distances along the centerline of the light source at a predetermined force level. 
A 5-pouch PPSA was suspended by two pieces of Spectra fiber (Power Pro 80 lb test). One end of the fiber was pinned and the other was
attached to a force gauge (Mark-10 M3-5) on an adjustable stage (Fig \ref{fig:outbox_schematic}A).

The PPSA was backed with a \SI{5}{\centi\meter} diameter LDPE tubing to simulate the heat transfer effects of the central spine. In each test, the system was equilibrated for 12 minutes to reach steady state and the displacement required to produce  \SI{5}{\newton} of force was measured. This allowed the contraction ratios to be measured as a function of distance (Fig. \ref{fig:SimpleEnvironment}C). Using experimental and theoretical arguments (see Sec. \ref{sec:ForceCompressionRatio}), we fitted a sigmoidal function \textcolor{black}{of the form $\gamma=c_1/(1+\exp(-c_2(x-c_3)))$ to the data. $c_1$, $c_2$, and $c_3$ are respectively, \SI{.209}{}, \SI{-24.145}{\meter^{-1}}, and  \SI{0.699}{\meter}.}

The uncertainty comes from the discrete adjustments of the stage used. The pin was moved between pre-drilled holes on the stage. We used a linear interpolation to find the contraction ratio that exerts \SI{5}{\newton}.
This experiment was designed to ensure that the illumination conditions were similar to those of the previous experiment, allowing us to estimate the relationship between contraction ratio and flux (Fig. \ref{fig:SimpleEnvironment}D). The exponential fit of Fig. \ref{fig:SimpleEnvironment}A was used to convert distance to flux of the fitted relationship in Fig. \ref{fig:SimpleEnvironment}C. \textcolor{black}{The relationship has the form of $\gamma=c_1/(1+\exp(-c_2(Q-c_3)))$ to the data. $c_1$, $c_2$, and $c_3$ are respectively, \SI{.206}{}, \SI{.0571}{\meter^{2}\watt^{-1}}, and  \SI{161.737}{\watt\meter^{-2}}.}

As long as the flux is known (and in the correct flux range, wavelengths, etc.), the relationship in Fig. \ref{fig:SimpleEnvironment}D allows us to predict the contraction of the PPSAs in environments beyond the centerline of Fig. \ref{fig:SimpleEnvironment}B. 



\begin{figure}[tb]
    \centering
    \includegraphics[width=.8\columnwidth]{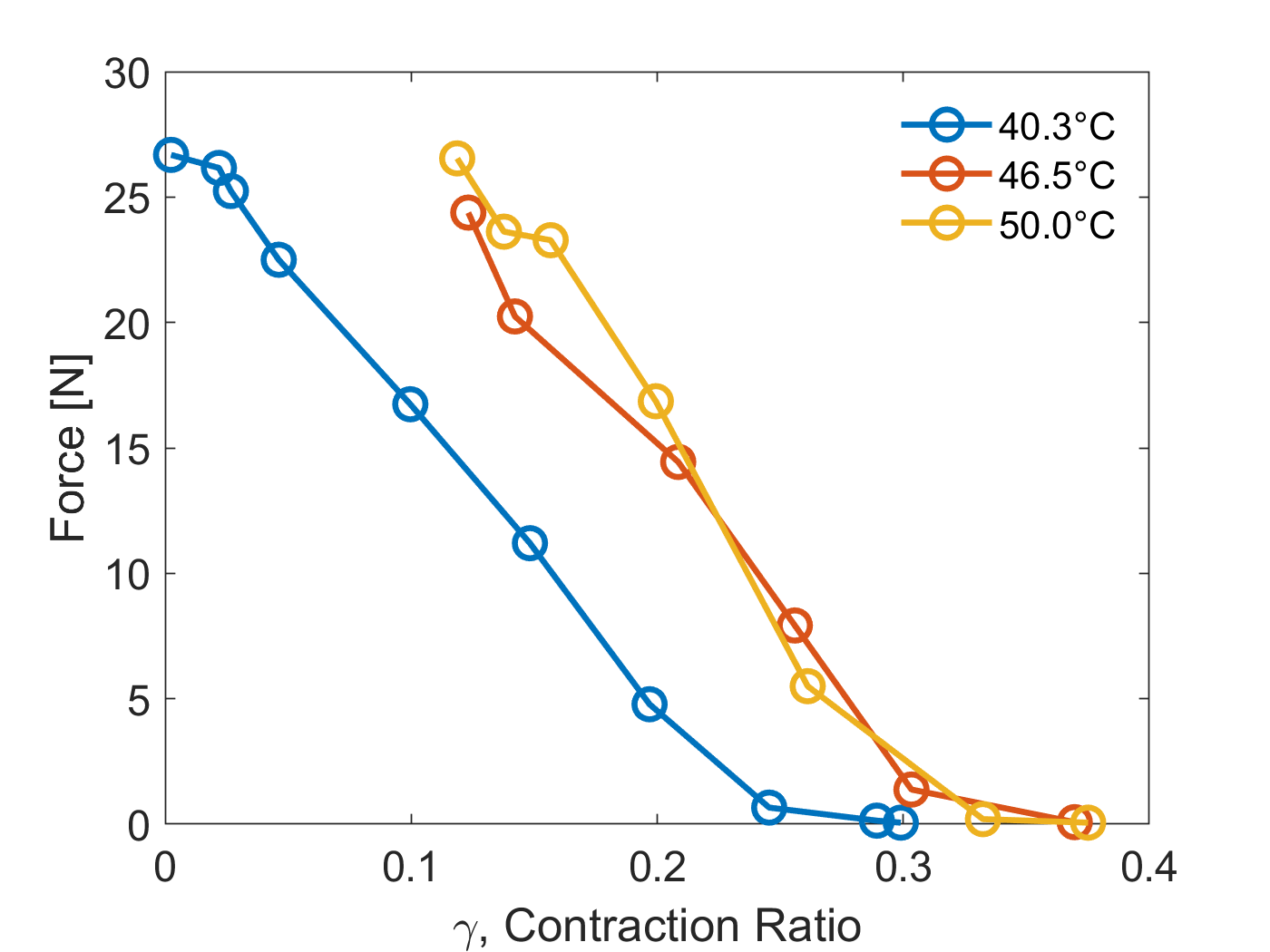}
    \caption{Force vs. contraction ratios for different temperatures. Moving from the lowest temperature (blue) to the middle temperature (red), the PPSA contracts substantially more at a given force level (e.g., \SI{5}{\newton}). However, a further increase in temperature (yellow) does not further increase contraction ratio. This behavior explains the sigmoidal shape in Fig. \ref{fig:SimpleEnvironment}C and D.}
    \label{fig:ForceStrain}
\end{figure}

\subsubsection{Interpretation}
\label{sec:ForceCompressionRatio}
The sigmoidal shapes in Fig. \ref{fig:SimpleEnvironment}C and D can be explained by the relationship between temperature and the shape of the PPSA force-contraction ratio curve (Fig. \ref{fig:ForceStrain}).  
To measure this, we placed PPSAs inside a temperature-controlled insulated box (Fig. \ref{fig:outbox_schematic}B). A linear stage adjusted the contraction ratio and a force gauge measured the forces.
We observed that as temperature (and thus internal pressure) increases, the location of the force-contraction ratio curve shifts to the right. Specifically, we observed that at a given force level, the contraction ratio initially increased as the temperature was increased from \SI{40.3}{\celsius} to \SI{46.5}{\celsius}. However, as the temperature was further increased to \SI{50.0}{\celsius}, saturation occurs, and the curves no longer shift to the right. This saturation effect with respect to temperature is reflected in the sigmoidal shapes seen in Fig. \ref{fig:SimpleEnvironment}C and D, \textcolor{black}{which show a plateau at $\sim$ \SI{250}{W/m^2} for our system}. 

We also note that while the previous models \cite{daerden1999conception,daerden2001concept} qualitatively describe the observed shapes in Fig. \ref{fig:ForceStrain} (i.e., a monotonic decrease in force with increasing contraction ratio), the models predict larger forces at a given contraction ratio and a larger maximum contraction ratio. This is possibly because the models did not consider the effects of materials with non-negligible bending stiffness and internal hysteresis. The \textcolor{black}{gas-impermeable} Mylar film we use is noticeably less supple than the polyethylene of \cite{greer2017series}.
As such, we instead used an empirical relationship for this component of our model.

\begin{figure}[tb]
    \centering
    \vspace{3mm}
    \includegraphics[width=.6\columnwidth]{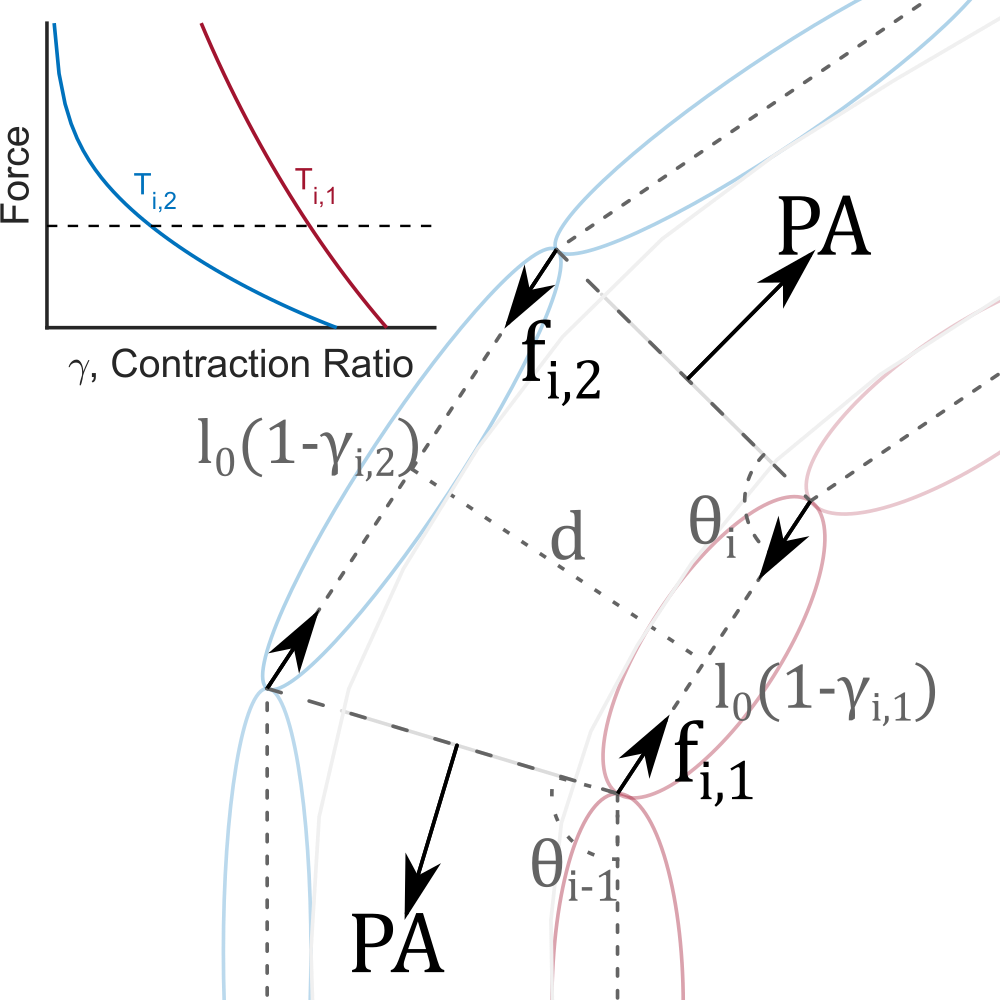}
    \caption{Detail of the static pose model. The outline of the robot is shown in light gray, the trapezoidal geometry and dimensions are shown in a darker gray, and the forces are shown in black. The higher temperature PPSA, which is closer to the light source, is shown in red (and its force-contraction ratio curve is shown on the inset). Similarly, the lower temperature PPSA is shown in blue. Note that the two PPSAs will have the same force, but different contraction ratios.}
    \label{fig:kinematicModel}
\end{figure}

\subsection{Analytical Component: Static Pose Model}
In this section, we present the analytical component of the model, which predicts the static pose of the robot given the force-contraction ratio relationship of each of the PPSAs along the robot body (determined from the empirical component of the model, given a light flux). The contraction-ratio of each PPSA varies due to temperature differences caused by differences in light flux at each PPSA. 

To form this static pose model, we first discretize the vine robot body into $N$ trapezoidal sections, each consisting of three springs (representative element shown in Fig. \ref{fig:kinematicModel}).
The center spring represents the pneumatic axial expansion force, and the two outer springs represent the PPSAs. We assume there are no forces between the trapezoidal sections (i.e., all forces are internal to each section), because we assume the robot is free from external contacts (i.e., horizontal on a frictionless surface).

The force and moment balances within the $i$-th section are respectively
\begin{equation}
\label{eq:F_balance}
    PA=f_{i,1}(\gamma_{i,1})+f_{i,2}(\gamma_{i,2}),
\end{equation}
\begin{equation}
\label{eq:M_balance}
    f_{i,1}(\gamma_{i,1})=f_{i,2}(\gamma_{i,2}).
\end{equation}
$P$ is the gauge pressure of the pneumatic backbone, $A$ is the cross-sectional area of the backbone, $f$ is the force exerted by the PPSA, and $\gamma$ is contraction ratio of the PPSA. The subscripts $1$ and $2$ denote which side of the robot the PPSAs are on. For Fig. \ref{fig:kinematicModel}, we chose the $1$ side to be the light source side and $2$ to be the side away from it.
For static balance to occur $f_{i,1}(\gamma_{i,1})=f_{i,2}(\gamma_{i,2})=PA/2$.  The $i$-th angle can be determined from the relation
\begin{equation}
    l_0(1-\gamma_{i,1})+\frac{d}{\tan(\pi-\theta_{i-1})}+\frac{d}{\tan(\pi-\theta_i)}= l_0(1-\gamma_{i,2}),
\end{equation}

where $d$ represents the width spacing between the two PPSAs. For simplicity, we assume this value is constant.

This model shows that even though the contraction ratio and temperature of opposing PPSAs may be different, the force exerted will be the same. This is because the tension in each of the two PPSAs at a given cross section will be equal and opposite to half of the axial expansion pressure inside the pneumatic backbone, as shown in \eqref{eq:F_balance} and \eqref{eq:M_balance}.

So at equilibrium, the force exerted by each PPSA depends solely on the pressure of the pneumatic backbone and is independent of temperature. However,  the exact contraction ratio of a certain PPSA depends on the pneumatic backbone pressure and the temperature of the PPSA. 
For example, if the backbone is exerting an expansion force of \SI{10}{\newton}, each of the two PPSAs on either side will be applying \SI{5}{\newton}, regardless of which one is closer to the light; however, the one that is closer to the light and hotter will have a greater contraction ratio (Fig. \ref{fig:kinematicModel}, Inset).

\section{Results}
\subsection{Robot Characterization and Model Verification}
\label{sec:verification}

\begin{figure}[tb]
    \centering
    \vspace{2.5 mm}
    \includegraphics[width=.65\columnwidth]{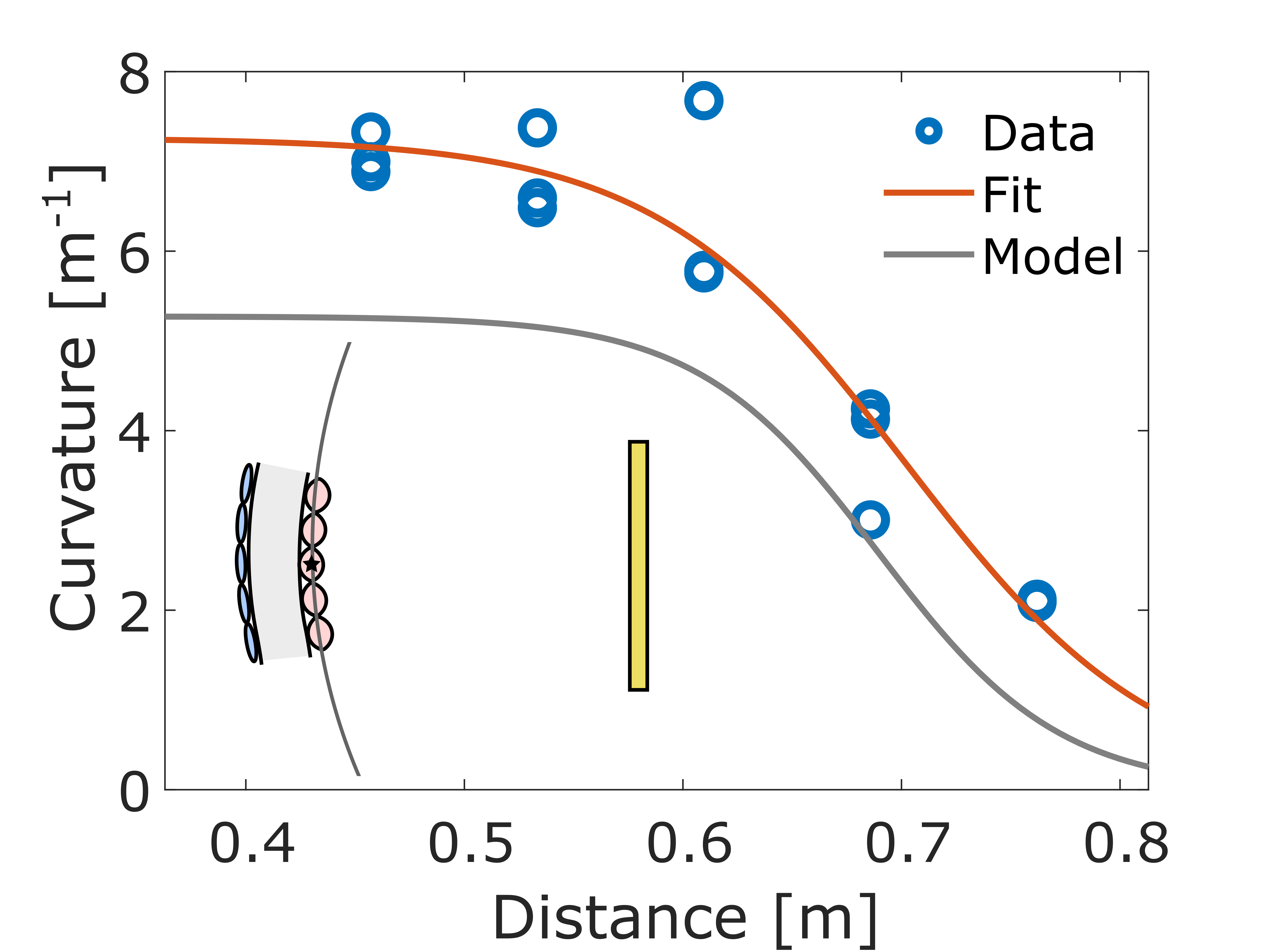}
    \caption{\textcolor{black}{Plot of the robot's curvature as a function of distance from the IR light. The measured curvature is larger than the predicted curvature, but still varies sigmoidally with distance. The inset is a schematic of the setup. Distance from the IR light (yellow bar) is measured from the point marked with a star. Curvature is measured from a fitted circle (grey) that passes through that point and the surrounding PPSAs' centers.}}
    \label{fig:curvature}
    \vspace{-5mm}
\end{figure}

\textcolor{black}{The first set of tests examined the actual and predicted curvature of the robot at different light flux levels, with the greatest tested approximately $<$40\% that of sunlight ($\sim$\SI{1000}{\watt}) (Fig. \ref{fig:curvature}). This served two purposes: characterization of the robot response and quantification of differences between the model and experiment.}

\textcolor{black}{To do so, we placed the robot at different distances from the IR light, with three trials for each distance.}
\textcolor{black}{We analyzed the center section of the robot
with illumination conditions similar to those described in Sec. \ref{sec:Contraction Ratio vs. Flux}.}
\textcolor{black}{The curvature was estimated by fitting a circle through the center points of the PPSAs that are closest to the IR light (opposite PPSAs were assumed to not contract).} 

\textcolor{black}{As shown in Fig. \ref{fig:curvature}, the experimentally measured curvature varied sigmoidally with distance. A sigmoidal curve fitted to the data suggests a maximum curvature of approximately \SI{7}{\meter^{-1}} at the highest flux values (i.e., closest distances). Knowing this is valuable for predicting paths the robot could feasibly take. We note that the measured curvature was fairly repeatable, with an average standard deviation for the different distances of \SI{0.5}{\meter^{-1}}}.
\textcolor{black}{
Similarly, the predicted curvature of the robot varied sigmoidally with distance. At close distances, the curvature saturated to about \SI{5.3}{\meter^{-1}} and at far distances it vanished. 
}
\textcolor{black}{The model, however, under-predicts the curvature. The fitted curve suggests a maximum curvature about a factor of about 1.4 larger than predicted. The under-prediction of the curvature of the robot can partially be explained by slight environmental and manufacturing variations between this test and the test used to generate Fig. \ref{fig:SimpleEnvironment}C.}

\vspace{-1.5mm}
\subsection{Light-seeking Behavior}
\begin{figure}[tb]
    \centering
    \vspace{2.5 mm}
    \includegraphics[width=.8\columnwidth]{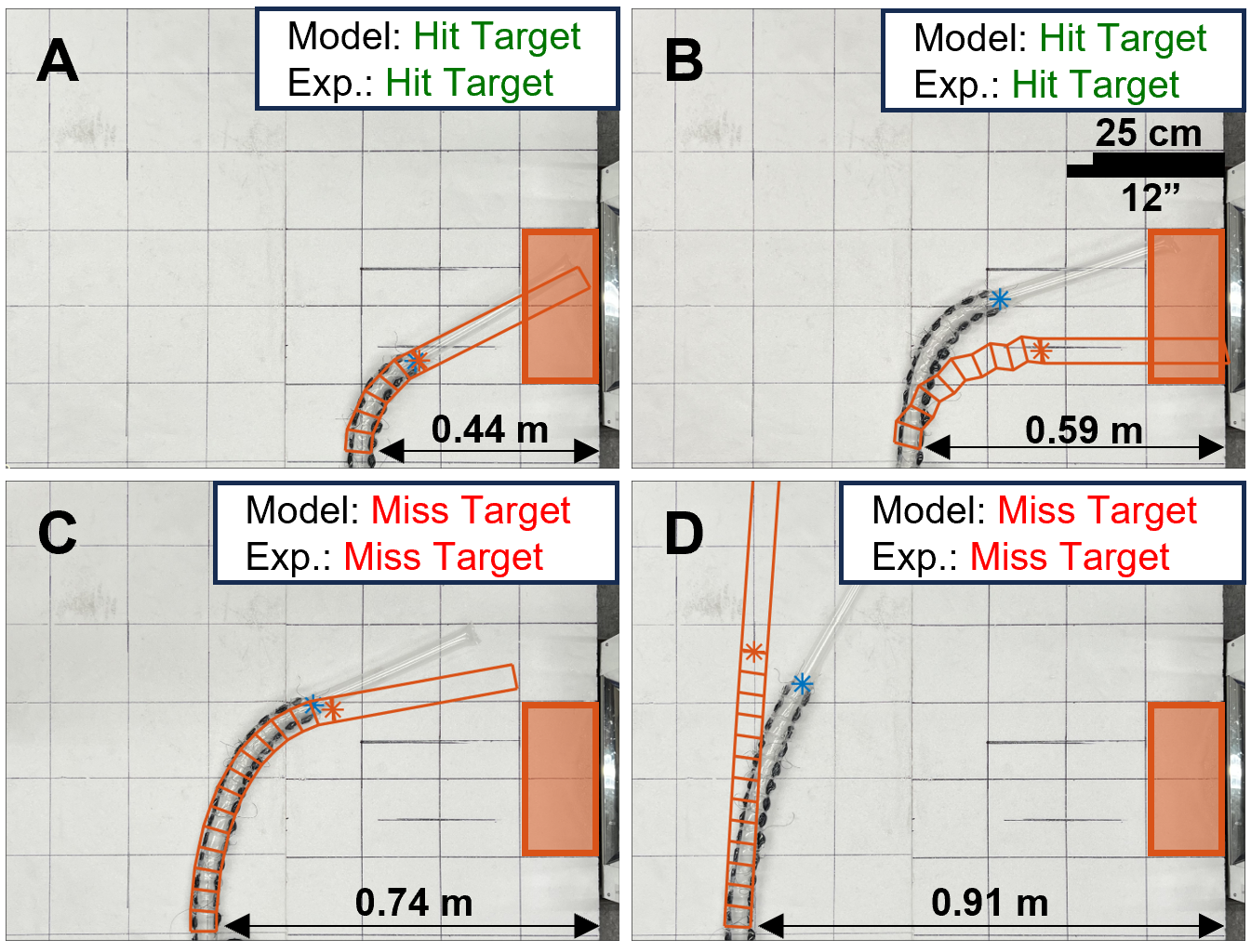}
    \caption{\textcolor{black}{Light-seeking behavior: the robot will find the light source when within \SI{0.59}{\meter}, and pass by if further than \SI{0.74}{\meter} away. The semi-empirical model can predict this in simple scenes. The target is the area directly in front of the light, and this area is overlaid with an orange rectangle. The predicted trajectories, in orange, are superimposed on the images. The test for case B ended before the robot reached equilibrium.}}
    \label{fig:HitTest}
    \vspace{-7.mm}
\end{figure}

\textcolor{black}{The second set of experiments tested the robot's ability to seek a light source as the initial position of the robot was changed with respect to the source location (Fig. \ref{fig:HitTest}). 
At the same time, we also compared the robot's behavior with that predicted by the model. 
In the future, a key use case of the model is predicting a robot's trajectory.} 

\textcolor{black}{For the test, we placed the robot at varying distances away from the IR light and let it move until it reached the target or stops moving.}
\textcolor{black}{
For the model, we use the same parameters as the curvature test. To account for the changing angle of the robot relative to the light source, we assume the backside only becomes active once it becomes parallel to the center axis of the light source, then it contracts the same amount as the active side.} 

\textcolor{black}{
As shown in Fig. \ref{fig:HitTest}, the robot reaches the target (the area directly in front of the IR light, marked in orange) when the starting offset from the light is less than \SI{0.59}{\meter}, and does not reach the target when offset more than \SI{0.74}{\meter}. Thus for this robot and a point light source of this power, if the robot grows within an approximately \SI{1.2}{\meter} circle around the source, it will bend and find the source. Sources with higher flux would have larger success circles.}

\textcolor{black}{Comparing to the model, we see that the model correctly predicts case A and B reaching the target and C and D not. We also evaluated its accuracy by measuring the distance between the experimental (blue) and model (orange) star markers. The star marks the end of the section with actuators. The error for A, B, C, and D are respectively \SI{0.01}{\meter}, \SI{0.13}{\meter}, \SI{0.04}{\meter}, and \SI{0.11}{\meter}. The test for case B ended before the robot reached equilibrium position; this partially explains the difference between the predicted and measured shape.}

\begin{figure}[tb]
\centering
\vspace{4 mm}
\includegraphics[width=0.8\columnwidth]{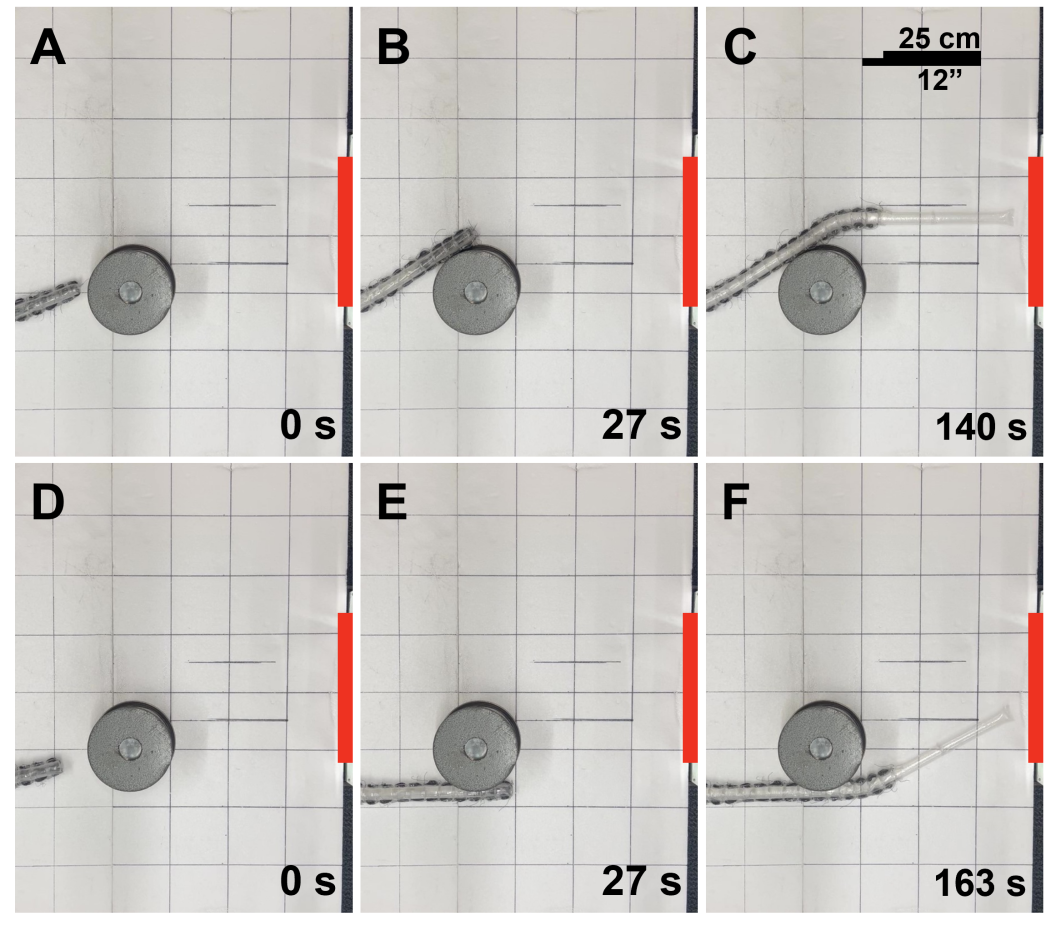}
\vspace{-5mm}
\caption{\textcolor{black}{The compliance of the robot allows it to deflect around an obstacle (circular object) and still grow towards a heat and light source. The vertical red bar marks the IR light location.}}
\label{fig:obTest}
\vspace{-5mm}
\end{figure}

\begin{figure}[tb]
    \centering
    \vspace{8 mm}
    \includegraphics[width=.6\columnwidth]{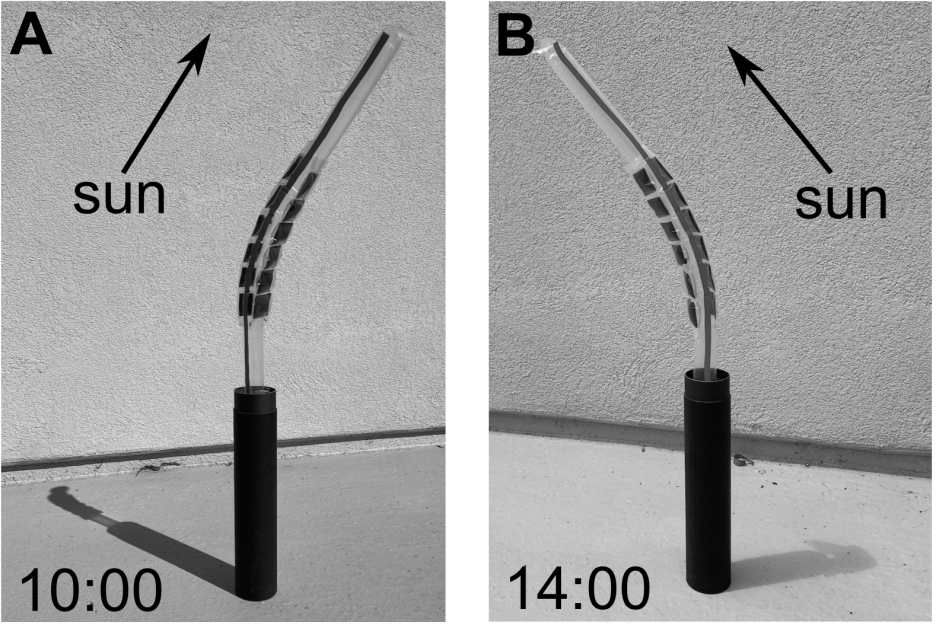}
    \caption{\textcolor{black}{Outdoor demonstration of robot responding to sunlight and pointing toward the sun. A: In the morning at 10:00, the robot bends to the right. B: In the afternoon at 14:00, it bends leftward.}}
    \label{fig:SunDemo}
    \vspace{-5mm}
\end{figure}

\subsection{Obstacle Interaction}
\textcolor{black}{Next, we tested the robot's ability to interact with obstacles in its environment. We grew the robot at an obstacle blocking its path, and observed the robot deflecting away from the original path, and then self-correcting back toward the light (Fig. \ref{fig:obTest}). When the robot is slightly offset to the left, it deforms toward the left and corrects to the right (Fig. \ref{fig:obTest}, A-C), and visa versa when offset to the right (Fig. \ref{fig:obTest}, D-F).}

\subsection{Sun Tracking}
\textcolor{black}{The PPSAs respond to light across a wide spectrum, meaning they respond to sunlight as well as the IR light used throughout other tests. To demonstrate this and the potential for the robot to eventually find applications in solar tracking, we placed the robot outside, facing vertically. It curved one direction toward the sun in the morning and the opposite direction in the afternoon (Fig. \ref{fig:SunDemo}).}

\subsection{Speed of Response}
\begin{figure}[h!]
\centering
\includegraphics[width=0.8\columnwidth]{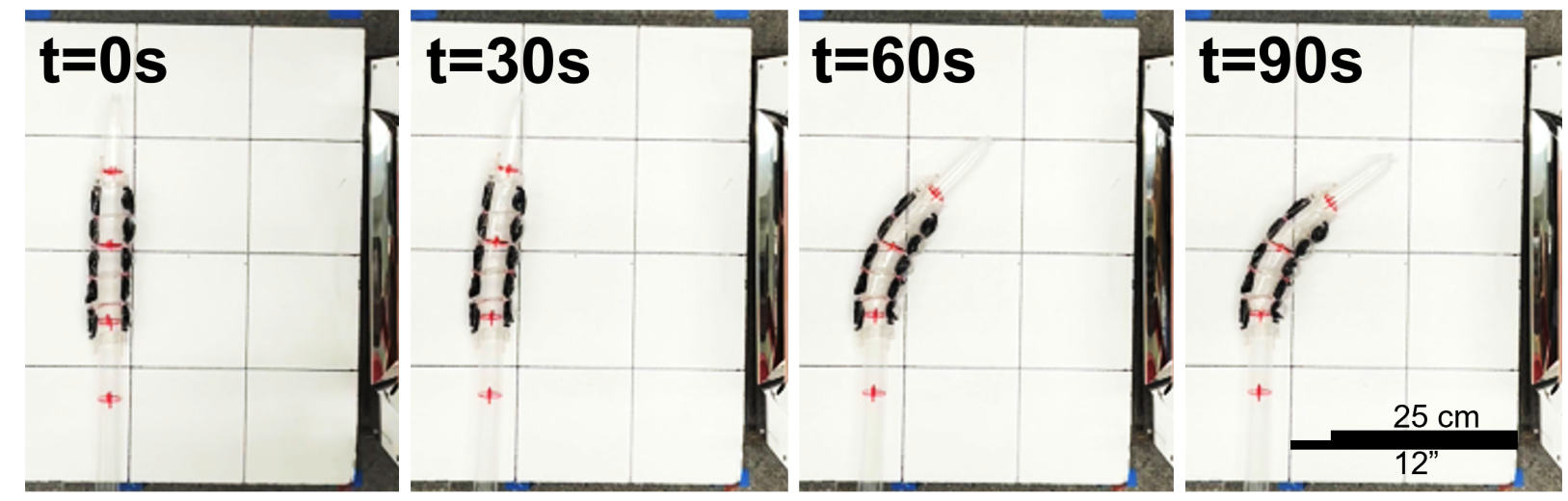}
\vspace{-3mm}
\caption{The robot can steer towards the heat source within \SI{90}{\second}.}
\label{fig:speedTest}
\vspace{-5mm}
\end{figure}

We quantified the response speed of the everted robot in the simple case of being placed next to a light source (at a distance of $\sim$\SI{35}{\centi\meter}) (Fig. \ref{fig:speedTest}). Full turning of the vine robot body by the photothermal PPSAs, each individually responding to the light, occurs in $\sim$\SI{90}{\second}. 

\begin{figure*}
\centering
\vspace{4mm}
\includegraphics[width=.8\linewidth]{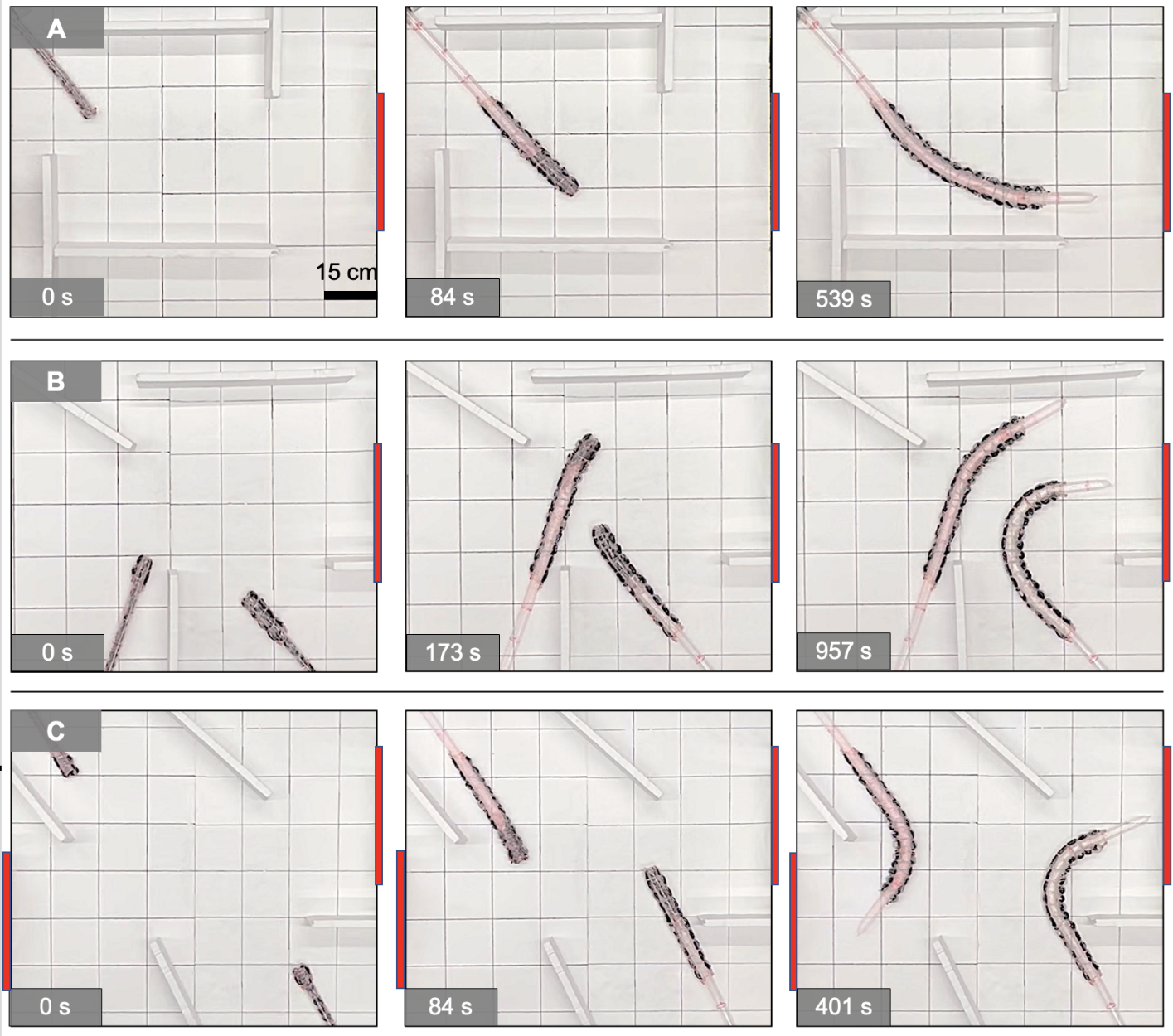}
\vspace{-3mm}
\caption{Demonstrations of light-seeking eversion toward the indicated light source(s) (represented by red boxes) in unknown environments. The PPSAs on opposing sides of the vine robot body actuate asymmetrically under photothermal activation, enabling the robot to bend and grow in a path defined by the styrofoam barriers. \textbf{A:} Eversion of one vine robot towards a light source on the right. \textbf{B:} Eversion of two vine robots towards the same light source. \textbf{C:} Eversion of two vine robots towards two light sources on the left and right. }
\label{fig:demo}
\vspace{-2mm}
\end{figure*}

\subsection{Variable Environment Testing}

Finally, we tested the robot in three different environments with obstacles between the robot(s) and light source(s). 
Fig. \ref{fig:demo}A shows the simplest case with a single robot and light source, and shows how the robot tip bends to grow straight towards the light source.
Fig. \ref{fig:demo}B adds a second robot, showing how these devices can work in parallel. Further, this shows with the right-hand robot that the robot can find the light source even if it is initially growing away from it from behind an obstacle. 
Fig. \ref{fig:demo}C shows another scenario with two robots growing in parallel, this time each seeking a different light source. This divergent behavior is enabled because each robot will respond most strongly to the light source that is closest to it.


\section{Conclusions and Discussion}

This work demonstrates the design, testing, modeling, and deployment of a light- and heat-seeking vine-inspired robot capable of autonomous motion toward a light or heat source. The robot makes use of a central pneumatically-pressurized spine for support and to drive eversion, while embedded material-level responsiveness in the PPSAs control direction of motion. We presented a semi-empirical model of the robot with sufficient resolution to capture the trends of the robot’s kinematic response to an external light source. This model was validated through experimental analysis of the pose in a simple IR light field. We quantified the ability of the robot to reach a target given different offsets, showed its ability to interact with obstacles and navigate around them, and demonstrated its ability to track the sun. 
Further, we showed multiple robots exhibiting light-seeking steering toward the same or different light sources. 

\vspace{-.75 mm}
\textcolor{black}{A natural application of the light- and heat-seeking robots is firefighting. While the presented concept is too slow and imprecise for fast-acting flaming fires such as house fires, it shows promise for slow acting smoldering (i.e., flameless combustion) fires, such as peat and coal seam fires and the remnants of flaming building and forest fires. These fires often persist long after the initial flaming fire and can last for years in some cases. For example, the debris of the World Trade Center burned for over 3 months \cite{Rein2016} and Burning Mountain in Australia, a coal seam fire, has been burning for about 6000 years \cite{smolderFire}.}

\textcolor{black}{Current firefighting methods for these fires are inefficient. Water is applied over large areas because the problem areas are difficult to locate and can occur in multiple areas. Additionally, subsurface structures such as soil pipes can divert water away from them. As a result, these fires need significant amounts of water to fully extinguish \cite{smolderFire}. }

\textcolor{black}{Climate change increases the importance of fighting these fires. Globally, peat and other organic soils store more carbon than the world's forests, and if burned, will release centuries of accumulated carbon into the atmosphere. Rising temperatures dry out these soils, making fires more likely \cite{smolderFire}; thus, improved firefighting methods are needed to mitigate potential climate impacts.}

\textcolor{black}{One potential solution is to have future versions of our vine robots navigate the natural pipe network in soils, the voids in rubble, or obstacles more generally, to find hot spots and deliver fire suppression agents. The more targeted approach might reduce the firefighting time significantly.}

\textcolor{black}{Obviously, the current iteration is far from ready. Firefighting versions will encounter temperatures up to \SI{700}{\celsius} and need to be much longer and agile and responsive enough to traverse through obstacle-strewn environments or peat pipe networks efficiently \cite{smolderFire}}.


Overall, this demonstration represents a significant step forward in our understanding of how to incorporate material-level responsiveness into vine-inspired robots and soft robots generally.


\section{Acknowledgements}
We thank \textcolor{black}{Riley Sandberg of Harvard University for suggesting the use of vine robots for peat fires} and David Haggerty for his editing help.




\bibliographystyle{IEEEtran}
\bibliography{IEEEabrv,Bibliography}

\end{document}